\documentclass[runningheads]{llncs}

\usepackage{eccv}

\usepackage{eccvabbrv}

\usepackage{graphicx}
\usepackage{booktabs}

\usepackage[accsupp]{axessibility}  % Improves PDF readability for those with disabilities.
\usepackage{microtype}
\usepackage[table]{xcolor}
\usepackage[breaklinks,colorlinks]{hyperref}

\newcommand{\red}[1]{{\color{red}#1}}
\newcommand{\todo}[1]{{\color{red}#1}}
\newcommand{\KG}[1]{{\color{magenta}#1}}

\newcommand{\TODO}[1]{\textbf{\color{red}[TODO: #1]}}
\newcommand{\KGnote}[1]{\textbf{\color{blue}[KG: #1]}}
\newcommand{\ASnote}[1]{\textbf{\color{orange}[AS: #1]}}
\newcommand{\KGcr}[1]{{\color{red}#1}}

\renewcommand{\KGnote}[1]{}
\renewcommand{\ASnote}[1]{}
\renewcommand{\KG}[1]{}
\renewcommand{\TODO}[1]{}
\renewcommand{\todo}[1]{}

\renewcommand{\red}[1]{}
\renewcommand{\KGcr}[1]{}

\usepackage[usenames,dvipsnames]{xcolor}
\hypersetup{
  colorlinks=true,
  citecolor=RoyalBlue,
  linkcolor=BrickRed,
  urlcolor=Blue
}

\usepackage{hyperref}

\usepackage{orcidlink}

\begin{document}

\title{ExpertEdit: Learning Skill-Aware\\Motion Editing from Expert Videos} 


\author{Arjun Somayazulu\inst{1}\orcidlink{0009-0005-1647-298X} \and
Kristen Grauman\inst{1}\orcidlink{0000-0002-9591-5873}}

\authorrunning{A.~Somayazulu, K. Grauman.}

\institute{The University of Texas at Austin, USA}

\maketitle

\begin{abstract}
Visual feedback is critical for motor skill acquisition in sports and rehabilitation, and psychological studies show that observing near-perfect versions of one's \emph{own} performance  accelerates  learning more effectively than watching expert demonstrations alone. We propose to enable such personalized feedback by automatically editing a person's motion to reflect higher skill. Existing motion editing approaches are poorly suited for this setting because they assume paired input-output data---rare and expensive to curate for skill-driven tasks---and explicit edit guidance at inference. We introduce \textbf{ExpertEdit}, a framework for skill-driven motion editing trained exclusively on \emph{unpaired} expert video demonstrations. ExpertEdit learns an expert motion prior with a masked language modeling objective that infills masked motion spans with expert-level refinements. At inference, novice motion is masked at skill-critical moments and projected into the learned expert manifold, producing localized skill improvements without paired supervision or manual edit guidance. Across eight diverse techniques and three sports from Ego-Exo4D~\cite{grauman2024egoexo4dunderstandingskilledhuman} and Karate Kyokushin~\cite{karate_dataset}, ExpertEdit outperforms state-of-the-art supervised motion editing methods on multiple metrics of motion realism and expert quality. Project page and benchmark available at: 
\url{https://vision.cs.utexas.edu/projects/expert_edit/}
\end{abstract}

\section{Introduction}
\label{sec:intro}

Imagine watching a video of yourself performing a layup, but with the smooth form, cadence, and precision of an NBA-level player. The shot still rises from your hands and arcs toward the basket, yet your body moves with trained efficiency, timing and rhythm, capturing the expert nuances of a player who’s practiced thousands of times. Just as Auto-Tune subtly corrects a singer’s pitch by a semitone while preserving their vocal timbre and phrasing, this imagined edit auto-tunes \textit{your motion}: infusing expert-like form into your exact execution while preserving your gross movement path and orientation in space. 

\begin{figure*}[t]
\centering
\includegraphics[width=0.7\linewidth]{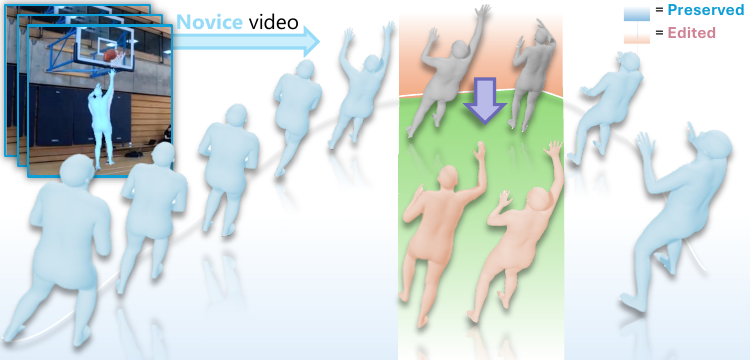}
\caption{\textbf{Skill-driven motion editing.} Given a 3D motion sequence extracted from a novice activity video, ExpertEdit produces personalized skill edits by refining poses within regions where skill differences are pronounced. Our model tweaks them to exhibit expert-like precision and form, while preserving the original execution's motion path and body orientation, as well as its source poses at all non-skill-critical moments. We learn to perform these edits solely from expert videos, without paired supervision or heavy edit guidance from text or reference motions---whether during inference or training. Preserved source poses  are shown in blue; edited poses are shown in orange.} 
\label{fig:concept_fig}
\vspace*{-0.15in}
\end{figure*}

Such personalized edited video has broad applications. Psychological research on video self-modeling shows that observing near-perfect versions of one’s \emph{own} performance improves motor skill acquisition, confidence, and retention~\cite{Steel2018, SteMarie2011, Ram2003}. Seeing one’s own body perform with expertise engages the perceptual–motor system more effectively than watching another person. This means personalized motion editing would be especially valuable for AI coaching, where individuals could receive feedback by viewing improved versions of their own movement within the original scene and viewpoint. Beyond coaching, such editing could help actors appear more skilled in stunts or choreography without body doubles, allow users to enhance their own performances (e.g., TikTok dances), and could even be used to refine the vast web of amateur activity footage into higher-quality human demonstrations for downstream robotics applications.

Despite this promise, skill-driven motion editing exposes a fundamental limitation of existing methods. Prior motion editing approaches assume either explicit edit conditioning—via text descriptions or reference clips~\cite{li2025simmotionedittextbasedhumanmotion, tu2024motionfollowereditingvideomotion, tu2023motioneditoreditingvideomotion, kim2023flamefreeformlanguagebasedmotion, jiang2025dynamicmotionblendingversatile, zhang2023finemogenfinegrainedspatiotemporalmotion}—or access to large-scale motion-text-motion triplets~\cite{delmas2024posefixcorrecting3dhuman, athanasiou2024motionfixtextdriven3dhuman, jiang2025dynamicmotionblendingversatile}. While these assumptions are reasonable for coarse edits (e.g., walking to dancing), they break down for skill refinement, where the goal is to improve execution of the same action rather than alter the action itself. Creating such supervision for skilled activities requires expert knowledge and significant manual effort. For example, MotionFix~\cite{athanasiou2024motionfixtextdriven3dhuman} constructs candidate motion pairs through large-scale motion similarity search followed by human annotation of edit descriptions, yet over 55\% of candidate pairs are rejected as `too similar' for labeling. Extending such pipelines to skill refinement would additionally require domain experts capable of diagnosing and describing subtle execution errors, making annotation difficult to scale. Consequently, no large-scale paired novice–expert motion dataset exists, and skill annotations remain scarce. At inference time, adapting the existing editing methods would require a human expert-in-the-loop to provide detailed text corrections or precisely aligned reference clips---precisely the form of feedback that AI coaching seeks to automate.

We therefore introduce \textbf{ExpertEdit} for \textit{skill-driven motion editing}: given a novice motion sequence, ExpertEdit transforms it into an expert-like performance (Fig.~\ref{fig:concept_fig}). Unlike traditional motion editing formulations, our goal is to improve execution of the same action---transforming a novice performance into an expert-like one---while preserving the actor’s identity and overall motion structure. Crucially, this process must operate without explicit edit guidance or reference demonstrations. In particular, we design ExpertEdit so it can be trained solely on unpaired expert motion and operate fully conditioning-free at inference time. By learning directly from expert demonstrations, our approach eliminates the need for paired supervision, expensive data curation, or explicit `correction' annotations.

Our key idea is to cast skill refinement as \textit{contextual motion infilling}. During training, ExpertEdit learns to reconstruct spans of motion using a masked language modeling (MLM) objective. Rather than masking arbitrarily, spans are centered around kinematic peaks---identified via simple motion statistics such as velocity or acceleration---where execution quality is most pronounced (e.g., takeoff in a layup). At test time, the same criteria select segments of a novice execution for infilling with expert-like refinements, effectively projecting key phases of novice motion onto the learned expert motion manifold.

Evaluated on eight diverse techniques spanning basketball, soccer, and karate across two datasets, ExpertEdit consistently outperforms state-of-the-art motion editing methods~\cite{li2025simmotionedittextbasedhumanmotion, athanasiou2024motionfixtextdriven3dhuman, kim2023flamefreeformlanguagebasedmotion} across multiple metrics of expert motion quality and realism, despite requiring no edit guidance at inference. These results establish skill-driven motion editing as a new capability for motion editing.

\vspace*{-0.05in}
\section{Related Work}
\vspace*{-0.05in}

\paragraph{Skill assessment from video.}
Skill understanding in video has been a longstanding challenge, with early benchmarks and methods for action quality assessment (AQA) in domains such as diving~\cite{xu2022finedivingfinegraineddatasetprocedureaware, parmar2019performedmultitasklearningapproach}, strength training~\cite{yin2025flexlargescalemultimodalmultiaction, parmar2022domain}, and figure skating~\cite{liu2024finegrainedactionanalysismultimodality}. Recent works expand this focus to multi-view, multi-agent sports including soccer~\cite{yeung2024autosoccerposeautomated3dposture,rao2025universalsoccervideounderstanding, grauman2024egoexo4dunderstandingskilledhuman, Noworolnik_2025_ICCV} and basketball~\cite{pan2025basketlargescalevideodataset, grauman2024egoexo4dunderstandingskilledhuman, wu2026skillsightefficientfirstpersonskill,ashutosh2026sportskillsphysicalskilllearning}. Early AQA works~\cite{parmar2017learningscoreolympicevents} treated skill as a regression task, while later efforts adopted ranking- and graph-based approaches for greater robustness~\cite{doughty2018whosbetterwhosbest}. More recent systems incorporate sub-action labels for procedure-aware evaluation~\cite{xu2024fineparserfinegrainedspatiotemporalaction}. Self-supervised approaches~\cite{parmar2022domain, 10.1145/3664647.3681084} learn skill-sensitive motion representations from unlabeled data using priors like periodicity and cycle-consistency. Recent work leverages narrations or assessment rubrics~\cite{zhang2024narrativeactionevaluationpromptguided, majeedi2024rica2rubricinformedcalibratedassessment, 10.1007/978-3-031-72946-1_24}.
However, these works focus on assessing or ranking skill rather than modifying motion to improve its execution quality. We address a generative problem: transforming novice motion into expert-like performance without paired supervision or explicit edit guidance.

\vspace*{-0.1in}
\paragraph{AI coaching from video.}
Recent work in AI coaching seeks to move beyond skill assessment by generating actionable feedback from video in the form of text commentary or reference clip retrieval. Several approaches generate corrective text suggestions based on pose deviations or visual analysis~\cite{yi2025exactvideolanguagebenchmarkexpert, ashutosh2025expertafexpertactionablefeedback, Yeh_2025, ashutosh2025learningskillattributestransferableassessment}. PoseTutor~\cite{9857137}, for example, highlights incorrect joint locations in static exercise images, relying on classifiers trained on labeled pose categories. Other works compute angular joint deviations between a source performance and an aligned expert reference to generate corrective text~\cite{Fieraru_2021_CVPR} or pose edits~\cite{Liu_2024, 10.1145/3746059.3747794}, requiring frame-level alignment with expert demonstrations. Retrieval-based systems such as Vid2Coach~\cite{huh2025vid2coachtransforminghowtovideos} provide step-level feedback for procedural tasks by leveraging large collections of visual and textual how-to resources. More recently, ExpertAF~\cite{ashutosh2025expertafexpertactionablefeedback} learns skill-corrective pose edits from paired novice–expert motion sequences annotated with expert commentary. In contrast to these approaches, which depend on text guidance, expert reference pairs, or supervision, ExpertEdit learns to refine motion directly from unpaired expert demonstrations. Our method generates skill-improved motion without requiring expert-provided commentary, frame-level alignment, or paired novice–expert samples during training. 

\vspace*{-0.1in}
\paragraph{Motion style transfer.}
Motion style transfer re-synthesizes a source motion sequence with stylistic characteristics derived from a reference motion. Classic approaches formulate this as a motion-style disentanglement problem using GANs~\cite{10.1145/3359566.3360083}, autoencoders~\cite{Aberman_2020, 10.1145/2820903.2820918, tao2022styleerdresponsivecoherentonline, Jang_2022}, or transformers~\cite{Kim_2024_CVPR}. Other work transfers motion across characters with different skeletons using kinematic constraints~\cite{villegas2018neuralkinematicnetworksunsupervised}, and diffusion-based approaches perform motion-to-motion translation driven by reference style signals~\cite{hu2024diffusionbasedhumanmotionstyle}. While effective for global attributes such as identity or affect, these approaches rely on synchronized reference motions at inference time. In contrast, ExpertEdit projects novice motion into the learned expert-style motion manifold at inference, refining novice execution without reference clips or explicit style signals.

\vspace*{-0.1in}

\paragraph{Instruction-driven motion editing.}
Text-driven motion editing has recently emerged as a popular paradigm. Recent methods apply transformer-based diffusion models to SMPL motion sequences, editing motions using low-level kinematic text prescriptions~\cite{delmas2024posefixcorrecting3dhuman, li2025simmotionedittextbasedhumanmotion}  (e.g., ``raise the right arm by 15 degrees") or discretized, code-level motion editing operators~\cite{Goel_2024}. Other works treat motion as a language modeling task, generating sequences autoregressively~\cite{zhang2023t2mgptgeneratinghumanmotion, lucas2022posegptquantizationbased3dhuman,jiang2025dynamicmotionblendingversatile}, with masked token modeling~\cite{guo2023momaskgenerativemaskedmodeling}, or jointly with text~\cite{jiang2023motiongpthumanmotionforeign,wang2024motiongpt2generalpurposemotionlanguagemodel}. 
Other motion editing approaches animate actor images or video clips conditioned on motion from pose sequences~\cite{chan2019everybodydance, liu2019liquidwarpingganunified, zhong2024decodecoupledhumancentereddiffusion, xu2023magicanimatetemporallyconsistenthuman, zhu2024champcontrollableconsistenthuman, hu2024animateanyoneconsistentcontrollable} or paired video clips~\cite{tu2023motioneditoreditingvideomotion, tu2024motionfollowereditingvideomotion, yang2020transmomoinvariancedrivenunsupervisedvideo}. While effective in human-in-the-loop settings, these approaches rely on privileged text or reference clips to guide the desired modifications. In contrast, ExpertEdit learns  skill-driven motion refinements by observing expert demonstrations, and automatically determines where and how a novice execution should be improved.

\vspace*{-0.1in}
\section{ExpertEdit}
\vspace*{-0.05in}

\begin{figure*}[t]
    \centering
    \includegraphics[width=1.0\linewidth]{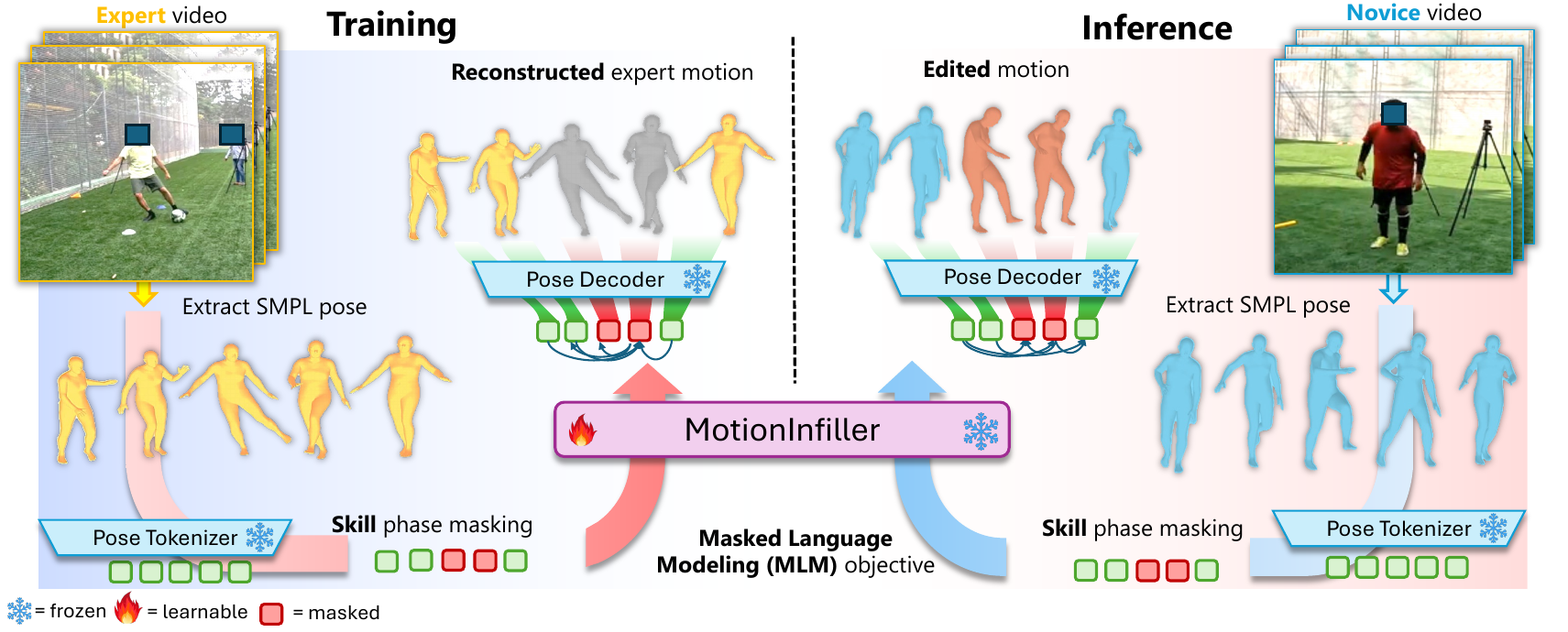}
    \caption{\textbf{ExpertEdit approach.} We tokenize expert pose motion sequences and mask the key action phase as determined by task-specific kinematic criteria. We train a bidirectional transformer, \textbf{MotionInfiller}, to predict the expert pose tokens at the masked positions. At inference, we mask skill-critical action phases in a novice motion (see Sec.~\ref{sec:experiments}) and infill these regions with expert-like motion.}
    \label{fig:model_fig}
    \vspace*{-0.1in}
\end{figure*}

We explore \textit{skill-driven motion editing}: the automatic transformation of a novice 3D motion sequence into an expert-like execution of the same action and duration, without text prompts, reference clips, or user-specified pose corrections.

Our goal is to synthesize body motion that exhibits expert-level precision and control while preserving the actor’s \textit{global translation}, \textit{root orientation}, and overall \textit{body shape}. We operate in 3D pose space, abstracting away scene context to focus on execution quality encoded in body kinematics. While scene context (e.g., distance to the hoop/net, position of the ball in basketball) influences an action, the quality of its execution is concentrated in body pose dynamics.

Rather than replacing the motion entirely, we perform temporally localized pose edits centered on kinematically defined phases of the action, leaving the remaining frames unchanged. We discuss this procedure later. The result is a smooth motion sequence that preserves the source execution's trajectory and identity, while reflecting expert-level form at key phases of the action where skill differences are pronounced.

\vspace*{-0.1in}
\subsection{Task formulation}
\label{sec:task}

We represent a motion sequence as 
\[
\mathbf{X} = \{ (\mathbf{r}_t, \mathbf{o}_t, \mathbf{p}_t) \}_{t=1}^{T},
\]
where $T$ is the sequence length, $\mathbf{r}_t \in \mathbb{R}^3$ denotes the global translation, $\mathbf{o}_t \in \mathbb{R}^3$ denotes the root orientation (in axis-angle form), and $\mathbf{p}_t \in \mathbb{R}^{3J}$ encodes the axis-angle rotations of $J$ skeletal joints at time $t$. The number of joints $J$ depends on the underlying pose representation (e.g., SMPL~\cite{SMPL:2015} or dataset-specific motion capture systems), but our formulation is agnostic to this choice.

Given a novice motion sequence $\mathbf{X}^{\text{nov}}$ demonstrating a particular skilled technique $\tau$, our goal is to generate an edited motion sequence $\mathbf{X}^{\text{edit}}$ of the same duration that reflects expert-level quality in the same execution. Importantly, we preserve the source execution’s global translation trajectory $\{\mathbf{r}_t\}_{t=1}^{T}$ and root orientation $\{\mathbf{o}_t\}_{t=1}^{T}$, restricting edits to the joint rotations $\mathbf{p}_t$ during selected motion phases. This ensures that the edited motion remains anchored to the original actor's motion path and movement structure while improving the skill level of the execution. To do this, we learn a mapping
\begin{equation}
\mathcal{F}_\theta^\tau : \mathbb{R}^{T \times 3J} \rightarrow \mathbb{R}^{T \times 3J}, 
\qquad 
\hat{\mathbf{p}}_{1:T} = \mathcal{F}_\theta^\tau(\mathbf{p}_{1:T}),
\end{equation}
where $\hat{\mathbf{p}}_t$ denotes the refined joint rotations for frame $t$, and $J$ is the number of skeletal joints. For each frame, we construct the edited motion as
\[
\mathbf{X}^{\text{edit}}_t = (\mathbf{r}_t, \mathbf{o}_t, \hat{\mathbf{p}}_t),
\]
where the root translation $\mathbf{r}_t$ and root orientation $\mathbf{o}_t$ are directly copied from the input sequence. This preserves the actor’s body shape, spatial trajectory, and orientation while allowing expert-like quality to emerge through learned body pose refinements across the motion sequence.

While temporal pacing and speed contribute to expertise, we preserve the input sequence’s duration and overall rhythm when editing motion, focusing the model on correcting joint coordination at skill-critical phases rather than altering the global tempo of the movement.

\paragraph{Technique criteria.}  ExpertEdit operates on action-centric motions corresponding to goal-directed athletic techniques such as a reverse layup, penalty kick, or a martial arts strike. These actions exhibit a relatively consistent execution structure across performers, making them well suited for learning expert motion priors. A key property of techniques is the presence of an \textit{operative moment}, a temporally localized event that determines the success of the action (e.g., ball release in a layup, or foot-ball contact in a penalty kick). The surrounding motion trajectory is organized around this moment, producing a canonical temporal structure shared across performers. We automate extraction of technique instances using text similarity between grounded narrations and technique operative moment descriptions (details in Supp.). In practice, technique instances can be identified within untrimmed recordings using grounded action annotations or narrations, or other temporal metadata commonly available in modern video datasets.

\vspace*{-0.05in}
\subsection{Approach}
\label{sec:approach}

We approach skill-driven motion editing as the problem of learning an \emph{expert motion manifold} from unpaired expert demonstrations, and using it to locally refine novice executions. Rather than generating motion from scratch, our goal is to project skill-critical phases of a novice sequence onto the manifold of expert motion while preserving the actor’s global trajectory, orientation, and body configuration. We achieve this through \textit{contextual motion infilling}: the model learns to reconstruct masked segments of expert motion during training, and at inference selectively replaces critical phases of a novice sequence %where skill differences emerge 
with expert-like refinements conditioned on surrounding motion context. This formulation enables localized, personalized skill enhancement without requiring paired novice–expert data, text instructions, or reference demonstrations.

At a high level, ExpertEdit has three steps: (1) learn a discrete expert-motion vocabulary, (2) train a bidirectional infiller to reconstruct kinematically selected expert motion spans, and (3) apply the same mask-and-infill procedure to novice motions at inference. See Fig.~\ref{fig:model_fig}.

We first train a \textbf{Pose Tokenizer} on expert motion sequences, learning a codebook of discrete, skilled motion tokens. We then train a bidirectional transformer, \textbf{MotionInfiller}, on tokenized expert sequences with a Masked Language Modeling (MLM) objective to infill kinematically selected windows of skill-critical moments conditioned on both past and future motion. At inference, we use the same kinematic criteria (defined below) to identify skill-critical moments in a novice sequence, and let the model infill those regions with expert-like poses accounting for the context of the surrounding novice poses. We retain the novice pose at unmasked regions, allowing the model to produce locally expert-like motion while retaining the actor's global trajectory and execution rhythm. Each component is described in detail below.

\paragraph{Pose tokenizer.}
We adopt a Transformer-based VQ-VAE~\cite{lucas2022posegptquantizationbased3dhuman} with causal self-attention as our \textbf{Pose Tokenizer}, where each encoded latent depends on the temporal history of preceding poses.  
Unlike frame-wise quantization, this formulation preserves temporal coherence and encodes motion dynamics directly into the latent sequence. Given expert motion sequence
\[
\mathbf{X}^{\text{exp}} = \{(\mathbf{r}^{\text{exp}}_t, \mathbf{o}^{\text{exp}}_t, \mathbf{p}^{\text{exp}}_t)\}_{t=1}^{T},
\]
we concatenate the root translation, root orientation, and joint rotations at each frame into a pose feature vector $
\mathbf{x}^{\text{exp}}_t = [\mathbf{r}^{\text{exp}}_t, \mathbf{o}^{\text{exp}}_t, \mathbf{p}^{\text{exp}}_t] \in \mathbb{R}^{6 + 3J}.$ The encoder \(E_\phi\) produces a sequence of latent vectors
\(\mathbf{Z} = [\mathbf{z}_{1}, \ldots, \mathbf{z}_{T}]\) where each latent
\(\mathbf{z}_{t}\) attends only to the motion up to time \(t\):
\begin{equation}
\mathbf{z}_{t} = E_\phi\big(\mathbf{x}^{\text{exp}}_{\leq t}\big)
= f_{\text{enc}}\!\big(\mathbf{x}^{\text{exp}}_{t},\, \text{Attn}_{\text{causal}}(\mathbf{x}^{\text{exp}}_{<t})\big),
\label{eq:vq_causal_encoding}
\end{equation}
where \(\text{Attn}_{\text{causal}}\) denotes transformer blocks with causal masking that restrict attention to prior frames. Each latent is then quantized to its nearest code in the learned codebook
\(\mathcal{E} = \{\mathbf{e}_{k}\}_{k=1}^{K}\):
\begin{equation}
k^{*}_{t} = \arg\min_{k}\ \|\mathbf{z}_{t} - \mathbf{e}_{k}\|_{2}^{2}, 
\qquad
\hat{\mathbf{x}}_{t} = D_\phi\big(\mathbf{e}_{k^{*}_{t}}\big),
\label{eq:vq_quant_causal}
\end{equation}
where \(D_\phi\) is a causal Transformer decoder that reconstructs each frame conditioned on the current and previously decoded tokens, $\hat{\mathbf{x}}_{t} = D_\phi(\mathbf{e}_{\leq t})$. The model is trained using the standard VQ-VAE loss:
\begin{equation}
\begin{split}
\mathcal{L}_{\text{VQ}} =
\|\mathbf{x}^{\text{exp}}_{t} - \hat{\mathbf{x}}_{t}\|_{2}^{2}
+ \|\text{sg}[E_\phi(\mathbf{x}^{\text{exp}}_{\leq t})] - \mathbf{e}_{k^{*}_{t}}\|_{2}^{2}
+ \beta\,\|E_\phi(\mathbf{x}^{\text{exp}}_{\leq t}) - \text{sg}[\mathbf{e}_{k^{*}_{t}}]\|_{2}^{2},
\label{eq:vq_loss_causal}
\end{split}
\end{equation}
where \(\text{sg}[\cdot]\) is the stop-gradient operator and \(\beta\) is the commitment weight. The tokenizer learns a discrete motion vocabulary in which each token encodes a short-term motion primitive, and a sequence of tokens is represented as $\mathbf{k} = [k_{1}, \ldots, k_{T}], 
\quad k_{t} \in \{1,\ldots,K\}$. The decoder \(D_\phi\) provides a geometry-consistent inverse mapping from tokenized motion back to full motion space.

\paragraph{Motion infilling.}
To perform context-conditioned motion editing, we adopt a \textbf{masked language modeling (MLM)} objective. Given tokenized expert motion sequence  
\(\mathbf{k}^{\text{exp}} = [k^{\text{exp}}_1, \ldots, k^{\text{exp}}_T]\), where \(k^{\text{exp}}_t \in \{1, \ldots, K\}\) during training, we use kinematic criteria to select a skill-critical motion peak index \(t^{*}\), described later.  

We then define a masked span length
\(\ell = \max(2, \lfloor \alpha T \rfloor)\),
where \(\alpha \in (0,1)\) controls the maximum proportion of the sequence to mask, and replace the corresponding token segment
\([k^{\text{exp}}_{t^{*}-h}, \ldots, k^{\text{exp}}_{t^{*}+h}],
\quad
h = \left\lfloor \frac{\ell}{2} \right\rfloor\)
centered at \(t^{*}\) with learned \texttt{[MASK]} tokens. This embedding is excluded from the output token vocabulary. The model is then trained to reconstruct the original tokens in the masked region given the unmasked temporal context on both sides:
\begin{equation}
    \mathcal{L}_{\text{MLM}}
    = - \sum_{i = t^{*}-h}^{t^{*}+h}
    \log p_\theta\!\big(k^{\text{exp}}_i \mid \mathbf{k}^{\text{exp}}_{\setminus [t^{*}-h:t^{*}+h]}\big),
    \label{eq:mlm_span_loss}
\end{equation}
where \(p_\theta(\cdot)\) is the transformer’s predicted probability over the token vocabulary of size \(K\).   
This cross-entropy loss encourages the network to infer contextually consistent motion that bridges masked spans with smooth and biomechanically valid transitions. We refer to this bidirectional transformer as \textbf{MotionInfiller}. 

ExpertEdit is technique-specific by design. For each technique $\tau$ (e.g., reverse layup, penalty kick), we train a separate Pose Tokenizer and MotionInfiller using the same architecture and objective. This lets each model learn a focused expert prior over a coherent motion family. This specialization is lightweight in our setting: adding a new skill requires only unpaired expert demonstrations, not paired novice--expert samples or per-instance text edits. Thus, technique-specific training gives the benefit of sharper expert priors without the annotation burden of constructing a paired, heterogeneous multi-skill correction dataset.

\paragraph{Kinematic phase selection.}
Skill differences in physical actions typically concentrate around a small number of biomechanically critical phases (e.g., takeoff in a basketball layup, ball contact in a soccer penalty kick). To discover these phases automatically, we compute a scalar kinematic signal \(h(t)\) over the motion sequence, derived from interpretable motion statistics such as vertical root velocity, joint acceleration, or jerk magnitude. The peak of this signal defines the skill-critical index
\[
t^{*} = \arg\max_{t \in \{1,\ldots,T\}} h(t).
\]
We then center the masked span at \(t^{*}\), as described earlier. The choice of kinematic signal is skill-specific but requires no paired supervision: for each technique, we select a motion statistic that consistently identifies the execution phase where skill differences are most pronounced, based on visual inspection of expert sequences (see implementation details in Sec.~\ref{sec:experiments}). 
Alternatively, one could automate this statistic choice, e.g., using LLM-proposed body-part signals per drill name, analogous to our LLM-derived operative phrases and extraction offsets (see Supp.). The same criterion is applied during both training and inference, ensuring that the model learns to refine precisely those phases where execution quality is most strongly expressed. 

\paragraph{Training and inference.}

For each technique, we first train a pose tokenizer on trimmed expert motion clips using Eq.~\eqref{eq:vq_loss_causal}. We then train MotionInfiller on tokenized expert sequences using the MLM objective in Eq.~\eqref{eq:mlm_span_loss}. At inference, we input a novice motion sequence \(\mathbf{X}^{\text{nov}}\) to the pose tokenizer \(E_{\phi}\) to produce a token sequence \(\mathbf{k}^{\text{nov}} = [k^{\text{nov}}_{1}, \ldots, k^{\text{nov}}_{T}]\). We then identify the skill-critical phase index \(t^{*}\) using the technique-specific kinematic criterion and replace the corresponding centered span with learned \texttt{[MASK]} embeddings. MotionInfiller one-shot infills expert motion tokens in the masked region, producing edited token sequence \(\hat{\mathbf{k}}\). Finally, \(D_\phi\) reconstructs the edited motion sequence \(\hat{\mathbf{X}} = \{(\mathbf{r}_t, \mathbf{o}_t, \hat{\mathbf{p}}_t)\}_{t=1}^{T}\) from \(\hat{\mathbf{k}}\), where root translation \(\mathbf{r}_t\) and orientation \(\mathbf{o}_t\) are copied from the input sequence and joint rotations have been edited (cf. Sec.~\ref{sec:approach}).

\vspace*{-0.1in}
 \section{Experiments}
 \vspace*{-0.05in}
\label{sec:experiments}

\paragraph{Datasets.} We evaluate ExpertEdit on eight diverse techniques across three sports---Basketball, Soccer, and Karate---from two datasets, Ego-Exo4D~\cite{grauman2024egoexo4dunderstandingskilledhuman} and Kyokushin Karate~\cite{karate_dataset}. We focus on sports datasets that capture wide variation in skill levels within a technique, enabling us to isolate expert executions for training and novice executions for evaluation.

\textbf{Ego-Exo4D~\cite{grauman2024egoexo4dunderstandingskilledhuman}} is a large-scale dataset of skilled human activity covering diverse physical activities and  skill levels, with timestamped, free-form narrations. Ego-Exo4D offers person-level proficiency score annotations in addition to grounded action narrations. We use "Late Expert" videos for our expert training set and "Novice" actor clips in the pairs for evaluation. We leverage all activities in Ego-Exo4D that exhibit a canonical motion path (see technique criteria, Sec.~\ref{sec:approach}): Mikan layup, Reverse layup, Mid-range jumpshot (basketball), and Penalty kick (soccer). We curate a training set of over 24k expert video clips (see Supp.) All motion sequences are extracted from exocentric views.

\textbf{Kyokushin Karate~\cite{karate_dataset}}  contains 1411 mocap recordings of 37 subjects performing four karate techniques: Reverse Punch, Spinning Back Kick, Front Kick, and Roundhouse Kick. The dataset contains 3229 single kicks and punches captured at 250 Hz. Each performer has a proficiency grade ranging from 9th kyu (least skilled) to 3rd dan (most skilled). We reclassify each sample as Expert (1st-3rd dan and 1st-3rd kyu) or Novice (6th-9th kyu). The train/val expert set is partitioned at the subject level to prevent the model from exploiting actor-specific motion idiosyncrasies.

\subsection{Constructing a paired test set for evaluation}
\label{sec:testset}

\begin{figure*}[t]
\centering
\includegraphics[width=\linewidth]{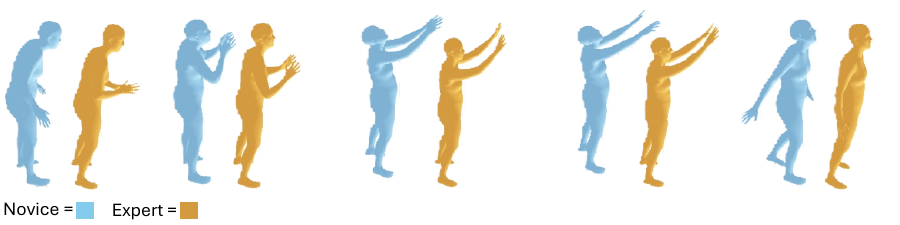}
\vspace*{-0.1in}
\caption{\textbf{Example novice--expert pseudo-pair sequence in our testbed.}  See text.
\vspace*{-0.15in}
}
\label{fig:pseudo_pair_fig}
\end{figure*}

To evaluate expert-like motion editing, we construct a high-quality testbed of temporally-aligned novice-expert pairs across both datasets. This allows us to compute frame-level motion quality metrics. Constructing skill-aligned pairs for evaluation follows recent precedent in prior coaching feedback work: ExpertAF~\cite{ashutosh2025expertafexpertactionablefeedback} similarly builds novice–expert video pairs (there for training). In contrast, we use manually verified novice--expert pairs only for evaluation.

For each trimmed novice clip, we retrieve $k=3$ expert clips within the same technique using similarity computed over available metadata (temporally annotated narrations e.g., "\textit{C shoots a layup into the net with his right hand}" for Ego-Exo4D and motion similarity for Kyokushin Karate). Pairing each novice with multiple expert references mitigates the effect of any single outlier pairing and captures natural variation in expert execution. While Karate techniques follow highly canonical forms such that pairs exhibit natural temporal alignment, basketball and soccer techniques exhibit greater variation in player distance to the ball or basket and starting position. To account for this variation, we temporally align each novice–expert pair using Dynamic Time Warping (DTW) on pose distance and resample the expert sequence to match the novice length. To address chirality asymmetries (e.g., opposite shooting hand), we compute alignment cost for both the original and sagittal-mirrored expert motion and select the lower-cost alignment.  

While we use these pairs only for evaluation, when fine-tuning supervised baselines, we separately construct a larger set of pseudo-pairs from the target datasets using the same retrieval procedure as above, but without expert verification or DTW alignment. This ensures baselines have target-domain paired exposure while avoiding expensive and unrealistically clean training samples--dense, human-verified, frame-aligned corrections--that ExpertEdit never receives.

\begin{figure*}[t]
\centering
\includegraphics[width=\linewidth]{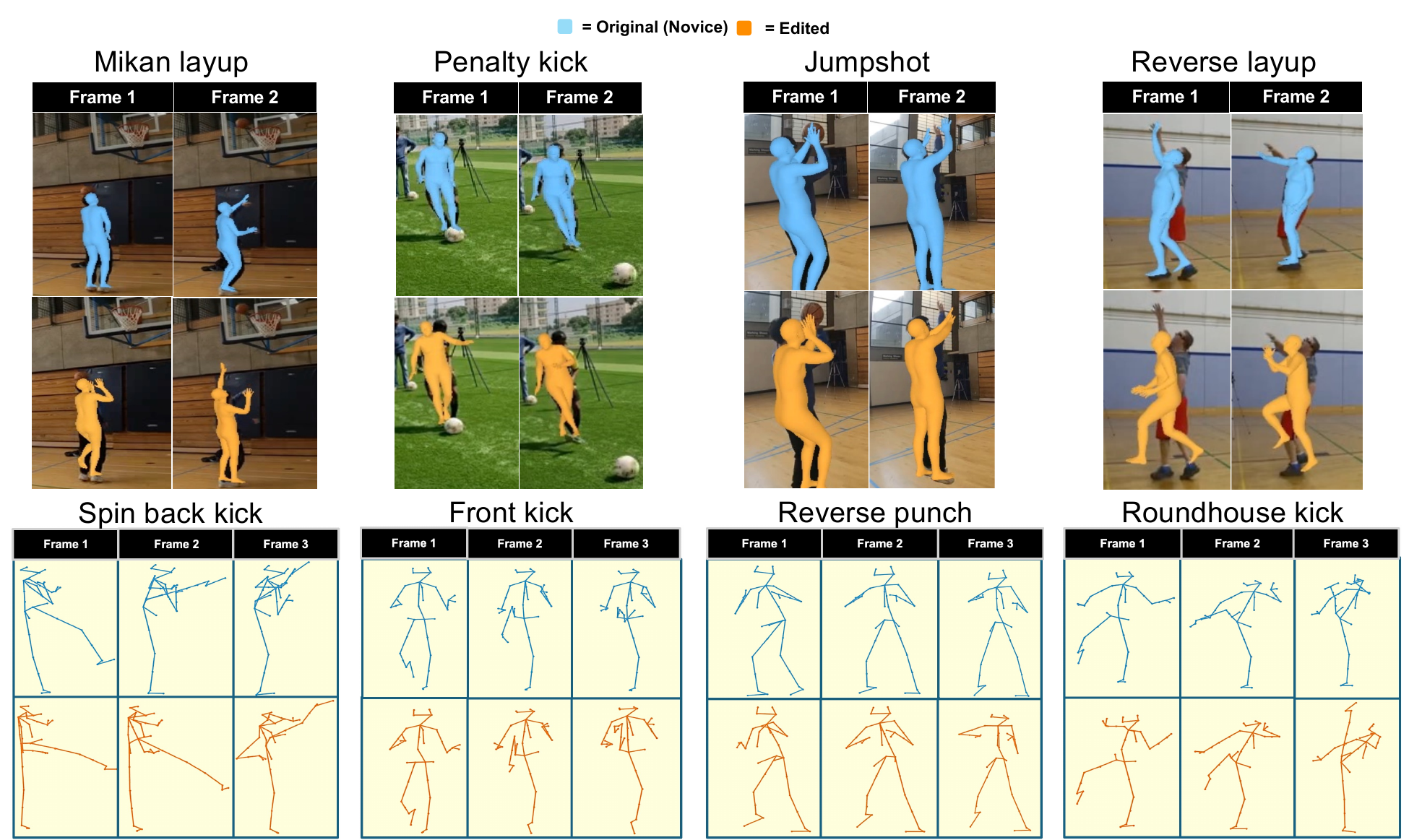}
\caption{\textbf{ExpertEdit sequence visualization:} We show novice source pose (blue) and edited pose (orange) at several frames for all techniques. ExpertEdit makes subtle pose refinements that improve form at skill-critical action moments, including raising the knee on the shooting hand-side higher during layups (Mikan, reverse), extending legs further on kicks (spin back, roundhouse), moving the shooting hand under the ball during jumpshots, and improving follow-through from the kicking leg on penalty kicks.}\vspace*{-0.15in}
\label{fig:vis_fig}
\end{figure*}

\vspace*{-0.1in}
\paragraph{Human validation of evaluation pairs.}
The pre-processed Karate Kyokushin dataset clips are trimmed at the atomic action level~\cite{Richardson_2025}, have formally defined proficiency ranks, and exhibit highly canonical forms, allowing us to form high-quality pairs from the existing trimmed clips. For our pseudo-pairs constructed with Ego-Exo4D, we recruit two basketball experts (28 combined years of playing experience) and two soccer experts (23 combined years) to independently assess the quality of the constructed pairs. Each rater evaluates whether (1) the novice and expert clips are temporally aligned, and (2) the expert clip clearly demonstrates higher skill and is correctly identifiable within the pair. A pair is retained if at least one rater judges it valid under both criteria. Using this protocol, 83\% of basketball pairs and 77\% of soccer pairs are deemed valid. These verified pairs constitute our final  evaluation test set, which is now publicly available and constitutes the first firm benchmark for this new task.

This procedure yields high-quality, temporally-aligned novice-expert evaluation pairs for each technique: Reverse layup (499), Penalty kick (173), Jumpshot (499), Mikan layup (436), Reverse punch (248), Spinning back kick (224), Front kick (231), and Roundhouse kick (239). 
Fig.~\ref{fig:pseudo_pair_fig} shows one such pair, a mid-range jumpshot. The expert follows the same technique and action-phase progression as the novice, while exhibiting expert-level differences in pose and form.

\paragraph{\textbf{Metrics.}} We evaluate whether generated motion exhibits expert-like characteristics using two expert-motion quality metrics. Metrics are computed only on frames within the kinematic-peak defined mask region for all methods.  We also provide \textbf{human-preference rates} as a quantitative metric of perceived quality.

\textbf{Pose Improvement ($\boldsymbol{P}$)} measures expert alignment using Procrustes-Aligned Mean Per-Joint Position Error (PA-MPJPE), a standard metric for evaluating 3D human pose accuracy~\cite{ashutosh2025expertafexpertactionablefeedback,yang2025posesynsynthesizingdiverse3d,yang2024egoposeformersimplebaselinestereo,zhao2026ppmotionphysicalperceptualfidelityevaluation,li2025genmogeneralistmodelhuman}. PA-MPJPE computes the distance between predicted and target 3D joint locations averaged over all joints and frames after scale and rigid alignment. We report relative improvement over the input novice motion:
\[
P(\%) = \frac{\text{PA-MPJPE}_{\text{novice}} - \text{PA-MPJPE}_{\text{gen}}}
{\text{PA-MPJPE}_{\text{novice}}} \times 100.
\]
Higher values indicate stronger movement toward expert-level execution. For each novice motion we generate $m=3$ edits and compute metrics against the $k=3$ paired experts to reduce reliance on any single expert trajectory. We report the minimum error across the $m$ generations.

\textbf{FID Improvement ($\boldsymbol{F}$)} measures whether generated motions align with the expert motion distribution using a Fréchet distance metric analogous to FID~\cite{heusel2018ganstrainedtimescaleupdate} used in generative image~\cite{hartwig2025surveyqualitymetricstexttoimage,karras2018progressivegrowinggansimproved,ho2020denoisingdiffusionprobabilisticmodels} and pose~\cite{lucas2022posegptquantizationbased3dhuman,zhang2023t2mgptgeneratinghumanmotion,zhang2022motiondiffusetextdrivenhumanmotion,tevet2022humanmotiondiffusionmodel,petrovich2022temosgeneratingdiversehuman,petrovich2021actionconditioned3dhumanmotion} evaluation. We train technique-level classifiers to distinguish novice from expert motions and use their intermediate feature representations to compute a Fréchet distance between generated motions and expert motions in feature space. As in Pose Improvement, we report relative FID improvement over the source novice distribution. DTW-based warping for frame-level alignment is only used when computing $\boldsymbol{P}$; $\boldsymbol{F}$ does not require frame-wise novice--expert correspondence.

\paragraph{\textbf{Baselines.}}
We evaluate ExpertEdit against several state-of-the-art motion editing methods, listed below. These approaches are designed for supervised motion editing and typically require paired demonstrations or explicit editing instructions during training. We fine-tune these baselines on a curated training set of 16k novice–expert pseudo-pairs derived using the same pairing procedure as our evaluation set, but without expensive frame-level alignment or human validation. These pairs constitute \emph{privileged supervision} provided only to the baselines: while ExpertEdit is trained exclusively on unpaired expert demonstrations, the baselines are allowed to learn directly from novice–expert pairs constructed from the target datasets. This gives the baselines training exposure to the target datasets for fair comparison. During both fine-tuning and inference, we use the text prompt ``Make the \texttt{[TECHNIQUE\_NAME]} motion smoother and more controlled''. Without access to expert-provided correctional text instructions for each novice sample, this prompt reflects a typical use case for skill-driven motion editing.  We arrived at this prompt after evaluating multiple prompts to elicit the best performance from the baselines; see Supp. for a prompt study. We found it produces more stable motion edits than prompts referring explicitly to “expert motion,” and more closely matches the style of text edits used in the datasets on which the baselines are pretrained~\cite{athanasiou2024motionfixtextdriven3dhuman,Guo_2022_CVPR}.

\begin{itemize}

\item \textbf{TMED~\cite{athanasiou2024motionfixtextdriven3dhuman}:} A diffusion transformer (DiT)–based motion editing model trained on MotionFix~\cite{athanasiou2024motionfixtextdriven3dhuman}. TMED performs text-conditioned motion editing by denoising motion tokens using cross-attention to the conditioning prompt.

\item \textbf{SimMotionEdit~\cite{li2025simmotionedittextbasedhumanmotion}:} A text-conditioned motion editing model that extends the TMED architecture with an auxiliary temporal guidance module. In addition to generating edited motion, SimMotionEdit predicts a 1D temporal signal indicating where edits should be applied, allowing the model to focus modifications on frames most relevant to the text prompt.

\item \textbf{FLAME~\cite{kim2023flamefreeformlanguagebasedmotion}:} A diffusion-based model designed for inference-time motion editing. FLAME generates edited motion conditioned on a source motion sequence and a text instruction. Unlike TMED and SimMotionEdit, which we fine-tune on our pseudo-paired supervision, FLAME does \textit{not} support paired fine-tuning, and is evaluated strictly in its inference-time editing mode.

\end{itemize}

We fine-tune TMED and SimMotionEdit from MotionFix~\cite{athanasiou2024motionfixtextdriven3dhuman} pretrained checkpoints and evaluate FLAME from a HumanML3D~\cite{Guo_2022_CVPR} checkpoint for Ego-Exo4D techniques. %\cc{For all baselines, we use the authors' released implementations and checkpoints; adaptations are limited to dataset loaders, pose-format conversion, dataset-statistics normalization, clip-length handling, and prompt formatting, with architectures unchanged.} 
For the Kyokushin Karate dataset, we evaluate only TMED and SimMotionEdit. FLAME is not included in this setting because its pretrained model operates on SMPL-based pose representations, whereas the karate dataset uses a custom MoCap joint parameterization. Applying FLAME would therefore require retraining the model from scratch in a new representation, which falls outside the inference-time editing paradigm for which the method was designed.

\paragraph{\textbf{Implementation details.}}
We extract SMPL~\cite{SMPL:2015} parameters from Ego-Exo4D videos using WHAM~\cite{shin2024whamreconstructingworldgroundedhumans} at 30 fps. For Kyokushin karate, we use the provided axis-angle joint rotations ($J=39$). For Ego-Exo4D techniques, we train pose tokenizers on fixed-length clips of 120 frames (reverse layup, penalty kick) and 90 frames (mikan layup, jumpshot). Karate sequences are uniformly resampled to 64 frames to normalize action duration while preserving relative motion timing. Our pose tokenizer follows the Transformer VQ-VAE architecture of~\cite{lucas2022posegptquantizationbased3dhuman}. MotionInfiller uses a bidirectional Transformer trained with span masking with maximum span fraction $\alpha=0.3$. See Supp. for an ablation on $\alpha$. Kinematic peak selection uses vertical velocity for basketball, maximum jerk for soccer, postural extremeness for reverse punch, and foot acceleration for karate kicks. Additional architecture, preprocessing, and training details are provided in Supp.

\subsection{Results}

\begin{table*}[t]
\centering
\scriptsize
\setlength{\tabcolsep}{3.5pt}
\renewcommand{\arraystretch}{1.1}

\caption{
\textbf{Results on four basketball and soccer techniques}.
$P$ and $F$ represent relative improvement in PA-MPJPE and alignment with the expert distribution respectively over the source motion. Higher is better for both ($\uparrow$).*Indicates no access to paired supervision for training.
PA-MPJPE is averaged over $k=3$ expert reference pairs to account for natural variation in expert behavior. See Supp. for absolute metrics.
}

\begin{tabular}{l cc cc cc cc}
\toprule

& \multicolumn{8}{c}{\textbf{Basketball and Soccer}} \\

\cmidrule(lr){2-9}

& \multicolumn{2}{c}{\textbf{Mikan}}
& \multicolumn{2}{c}{\textbf{Rev. Layup}}
& \multicolumn{2}{c}{\textbf{Jumpshot}}
& \multicolumn{2}{c}{\textbf{Pen. Kick}} \\

\cmidrule(lr){2-3}
\cmidrule(lr){4-5}
\cmidrule(lr){6-7}
\cmidrule(lr){8-9}

\textbf{Method}
& $P(\uparrow$) & $F(\uparrow$)
& $P(\uparrow$) & $F(\uparrow$)
& $P(\uparrow$) & $F(\uparrow$)
& $P(\uparrow$) & $F(\uparrow$) \\

\midrule

SimMotionEdit~\cite{li2025simmotionedittextbasedhumanmotion}
& 2.26 & 1.28 & 
2.31 & 4.88 & 
3.95 &  1.24 & 
3.20 &  2.38  \\

TMED~\cite{athanasiou2024motionfixtextdriven3dhuman}
& 1.87 & 1.00 & 
2.20 & 4.30 & 
2.58 & 2.03 & 
2.57 &  1.95  \\

FLAME*~\cite{kim2023flamefreeformlanguagebasedmotion}
& 2.35 & 1.45 & 
2.09 & 6.13 & 
3.13 & 1.54 & 
2.90 &  2.22  \\

\rowcolor{gray!15}
\textbf{ExpertEdit* (Ours)}
& \textbf{6.18} & \textbf{5.72} & 
\textbf{5.87} &  \textbf{12.08} & 
\textbf{5.34} & \textbf{7.66} & 
\textbf{6.03} & \textbf{9.14}  \\

\bottomrule
\end{tabular}
\label{tab:egoexo}
\end{table*}

\paragraph{\textbf{Basketball and soccer techniques.}}
Results on basketball and soccer are shown in Table~\ref{tab:egoexo}. Despite lacking access to privileged supervision provided to SimMotionEdit~\cite{li2025simmotionedittextbasedhumanmotion} and TMED~\cite{athanasiou2024motionfixtextdriven3dhuman}, ExpertEdit outperforms state-of-the-art motion editing baselines across all four techniques, achieving roughly $2$--$4\times$ larger gains in pose improvement ($P$) and expert FID score improvement ($F$) compared to the strongest baselines. The largest gains occur for reverse layups and penalty kicks, where ExpertEdit achieves $F=12.08\%$ and $9.14\%$ respectively, compared to at most $6.13\%$ and $2.38\%$ for competing approaches.

Fig.~\ref{fig:vis_fig} (top row) illustrates these corrections on sample edited frames. On the reverse layup (right), the edited motion lifts the shooting side knee higher during the jump and release phase, while in penalty kicks (middle left) the model exaggerates follow through on the striking leg after ball contact. These localized corrections lead to motions that better match both expert pose trajectories and the overall expert motion distribution.

These gains reflect the fact that skill differences often manifest in short, temporally localized moments of an action. While SimMotionEdit predicts a temporal signal to identify frames requiring stronger edits, it relies on descriptive text prompts to determine this signal and how the motion should change, which can suffer under the general text edit prompt required in this setting. In contrast, ExpertEdit targets skill-critical motion phases through kinematic masking identified from the source motion itself, allowing the model to focus edits on the moments where expert execution differs most strongly.

\begin{table*}[t]
\centering
\scriptsize
\setlength{\tabcolsep}{3pt}
\renewcommand{\arraystretch}{1.05}

\caption{
\textbf{Results on Kyokushin Karate Dataset, four techniques}.
$P$ denotes relative improvement in PA-MPJPE.
$F$ denotes relative improvement in FID over novice motion.
Higher is better for both ($\uparrow$).
}
\begin{tabular}{l cc cc cc cc}
\toprule

& \multicolumn{8}{c}{\textbf{Kyokushin Karate}} \\

\cmidrule(lr){2-9}

& \multicolumn{2}{c}{\textbf{Rev. Punch}}
& \multicolumn{2}{c}{\textbf{Spin BK}}
& \multicolumn{2}{c}{\textbf{Front}}
& \multicolumn{2}{c}{\textbf{High RH}} \\

\cmidrule(lr){2-3}
\cmidrule(lr){4-5}
\cmidrule(lr){6-7}
\cmidrule(lr){8-9}

\textbf{Method}
& $P(\uparrow$) & $F(\uparrow$)
& $P(\uparrow$) & $F(\uparrow$)
& $P(\uparrow$) & $F(\uparrow$)
& $P(\uparrow$) & $F(\uparrow$) \\

\midrule

SimMotionEdit~\cite{li2025simmotionedittextbasedhumanmotion}
& \textbf{2.20} & 1.30
& 1.43 & 3.30
& 1.45 & 5.05
& \textbf{3.30} & 5.88 \\

TMED~\cite{athanasiou2024motionfixtextdriven3dhuman}
& 0.92 & 1.08
& 0.86 & 2.25
& 0.38 & 2.90
& 2.25 & 3.37 \\

\rowcolor{gray!15}
\textbf{ExpertEdit (Ours)}
& 2.07 & \textbf{1.36}
& \textbf{1.79} & \textbf{4.23}
& \textbf{1.88} & \textbf{9.73}
& 2.96 & \textbf{6.18} \\

\bottomrule
\end{tabular}
\label{tab:karate}
\end{table*}

\vspace*{-0.1in}
\paragraph{\textbf{Kyokushin karate techniques.}}
Results on the Kyokushin Karate techniques are shown in Table~\ref{tab:karate}. ExpertEdit consistently improves alignment with the expert motion distribution, achieving the largest gain in FID Improvement ($F$) across all techniques. While SimMotionEdit achieves slightly higher Pose Improvement ($P$) on two techniques (Rev. Punch and High Roundhouse), these gains do not translate into stronger expert distribution alignment: ExpertEdit still achieves higher $F$ scores on both techniques. This suggests that while supervised models may produce edits that reduce pose error relative to individual expert examples, they do not necessarily generate motions that better match the overall expert motion distribution. While the absolute scale of improvements is smaller than in the basketball and soccer scenarios, this reflects characteristics of the dataset: karate techniques follow highly canonical motion patterns, and our expert pool includes both dan practitioners and upper-level kyu ranks (1st–3rd kyu) to ensure sufficient training data. This produces a narrower skill gap between actors and a less sharply defined expert motion manifold.

Fig.~\ref{fig:vis_fig} (bottom row) illustrates our results on karate techniques. Across multiple kick techniques, the edited motion (orange) consistently produces better leg extension than the source clip. These corrections correspond to the large improvements in expert distribution alignment observed for these techniques.   See Supp. video for examples with motion.

\vspace*{-0.1in}
\paragraph{\textbf{Human preference study.}}
Finally, to evaluate whether generated edits are perceived as useful visual feedback, we conduct a blinded A/B preference study on rendered edits. We focus this study on the basketball and soccer domains, where we could recruit target users with relevant playing experience to judge form improvements. For each trial, participants view the source motion and two anonymized edited motions from ExpertEdit and a strong fine-tuned baseline, SimMotionEdit~\cite{li2025simmotionedittextbasedhumanmotion}, with method order randomized. We ask: ``Which edited motion would be more useful to improve your form if you were the performer in the video?'' Participants choose from ``A'', ``B'', or ``Neither''.

Table~\ref{tab:user_study} summarizes the results. Across 24 raters and 112 total pairwise judgments, ExpertEdit is preferred in 49 judgments versus 29 for SimMotionEdit, with 34 ``Neither'' responses. Among non-neutral judgments, ExpertEdit is preferred 62.8\% of the time, suggesting that our localized and skill-driven edits are perceived as more useful corrective feedback than the compared text-driven motion editor. These results augment the metrics reported above by directly measuring perceived feedback usefulness. 

\begin{table}[t]
\centering
\scriptsize
\setlength{\tabcolsep}{4pt}
\caption{\textbf{Blinded human preference study.} Participants compare anonymized ExpertEdit and SimMotionEdit (strongest baseline) rendered edits and choose which would be more useful for improving form.}
\label{tab:user_study}
\begin{tabular}{lccc}
\toprule
Preferred edit & Basketball (78) & Soccer (34) & Overall (112) \\
\midrule
ExpertEdit & \textbf{42.3\%} & \textbf{47.1\%} & \textbf{43.8\%} \\
SimMotionEdit~\cite{li2025simmotionedittextbasedhumanmotion} & 26.9\% & 23.5\% & 25.9\% \\
Neither & 30.8\% & 29.4\% & 30.4\% \\
\midrule
ExpertEdit among non-neutral & \textbf{61.1\%} & \textbf{66.7\%} & \textbf{62.8\%} \\
\bottomrule
\end{tabular}
\end{table}

\vspace*{-0.05in}
\section{Conclusion}
\label{sec:conclusion}
\vspace*{-0.05in}

We introduce \textbf{ExpertEdit}, a framework for skill-driven motion editing that refines novice demonstrations into expert-like executions by projecting skill-critical phases in a novice motion onto a learned manifold of expert motion. Unlike prior approaches that rely on text guidance, reference clips, or heavy paired supervision, ExpertEdit learns an expert motion prior directly from unpaired expert performances and applies it to novice inputs at inference without additional conditioning. Across diverse techniques spanning basketball, soccer, and karate from two datasets, ExpertEdit outperforms state-of-the-art motion editing methods on multiple expert-quality metrics, including supervised approaches that rely on privileged paired demonstrations. These results establish skill-driven motion refinement as a previously unmet capability in motion editing.  By forgoing %requiring 
expensive skill-annotated motion pairs, 
ExpertEdit unlocks the 
%Overall, ExpertEdit learns effective expert motion priors even in a lower-data regime with different pose representations and smaller skill gaps. The ability to improve expert motion alignment without paired supervision highlights 
potential to tackle diverse motion domains simply by watching how experts perform.

\section*{Acknowledgements}

This research is supported in part by the UT Austin IFML
NSF AI Institute, Lyda Hill Philanthropies, and a gift from Amazon. We thank Ashutosh Kumar and Joungbin An for helpful discussions.

\bibliographystyle{splncs04}
\bibliography{main}

% ---------------------------------------------------------------
% Supplementary formatting

% Change section numbering to letters: A, B, C, ...
\renewcommand{\thesection}{\Alph{section}}
\renewcommand{\thesubsection}{\thesection.\arabic{subsection}}
\renewcommand{\thesubsubsection}{\thesubsection.\arabic{subsubsection}}

\title{ExpertEdit: Learning Skill-Aware\\Motion Editing from Expert Videos\\[0.5em]
\small \textmd{Supplementary Material}}

\titlerunning{ExpertEdit Supplementary Material}

% Keep anonymous for review
\author{}
\authorrunning{}

\institute{}

\maketitle

\begin{table}[h]
\centering
\normalsize
\begin{tabular}{p{0.9\linewidth}r}
\textbf{Table of Contents} & \\
\toprule
1) Dataset preprocessing and technique clip extraction pipeline  \dotfill & Sec.~\ref{sec:techniques} \\

2) Details of the procedure for constructing test set pseudo-pairs \dotfill & Sec.~\ref{sec:eval_pair} \\

3) Additional model architecture and training details \dotfill & Sec.~\ref{sec:imp_det_full} \\

4) Performance vs. train set size scaling analysis \dotfill & Sec.~\ref{sec:scaling_analysis} \\

5) Experiments with different text prompts for baselines \dotfill & Sec.~\ref{sec:prompt_experiments} \\

6) Ablation on mask span fraction $\alpha$ \dotfill & Sec.~\ref{sec:mask_span_ablation} \\

7) Condensed absolute metrics table \dotfill & Sec.~\ref{sec:absolute_metrics_sec} \\

8) Supplementary video content overview \dotfill & Sec.~\ref{sec:supp_video} \\
\bottomrule
\end{tabular}
\end{table}

\section{Dataset preprocessing and technique clip extraction}
\label{sec:techniques}

We describe our procedure for extracting technique-centered clips from Ego-Exo4D (c.f.~Sec. 3.1 `Technique criteria', ~Sec. 4 `Datasets'). For Kyokushin karate, we directly use the pre-trimmed clips provided by MoDiffAE~\cite{Richardson_2025}.

\subsection{Operative moment mining} 
Given an untrimmed expert video $E$ of technique $\tau$
and associated set of action narrations
$\mathcal{N}_{E}=\{(t_k, s_k)\}_{k=1}^{K_E}$,
where $t_k$ is a single timestamp and $s_k\in\mathcal{L}$ is a free-form caption describing a fine-grained action, we prompt an LLM with the technique description to produce a set of operative phrases
$\Phi_{\tau}=\{\phi_i\}_{i=1}^{m_\tau}$ (e.g., \textit{“ball leaves hand},” \textit{“ball contacts foot},” \textit{“shot released”}). We generate the operative phrase set once per technique and keep it fixed across all videos of that technique. We score narration–phrase affinity using a sentence encoder $f_{\text{enc}}:\mathcal{L}\to\mathbb{R}^d$:
\begin{equation}
\begin{split}
s_{i,k}(E)=\cos\!\left(f_{\text{enc}}(\phi_i),\ f_{\text{enc}}(s_k)\right),\qquad (\phi_i\in\Phi_\tau,\ (t_k,s_k)\in\mathcal{N}_{E}),
\label{eq:sem_sim}
\end{split}
\end{equation}
and retain matches above a threshold $\theta_{\text{sim}}=0.5$.

\subsection{Technique-specific extraction windows} Different techniques exhibit distinct temporal asymmetries between the buildup, execution, and follow-through phases of motion. To capture the full motion surrounding each operative moment, we use per-technique temporal offsets $\delta^{-}_{\tau}$ and $\delta^{+}_{\tau}$ denoting the number of frames to include before and after the operative moment. These are estimated once from the technique description using an LLM prior to dataset construction and held fixed across all clips.

\begin{equation}
\mathbf{X}^{e}_{1:T} = 
\text{extract}\!\left(t^\star - \delta^{-}_{\tau},\ 
t^\star + \delta^{+}_{\tau}\right),
\qquad
T = \delta^{-}_{\tau} + \delta^{+}_{\tau} + 1,
\label{eq:extract_train}
\end{equation}
where $\tau$ indexes the technique category, ensuring that each clip consistently captures the full canonical motion sequence for that technique. We use this procedure to extract trimmed technique instances from both expert videos for training as well as novice videos for our paired evaluation set.

\section{Evaluation pair construction}
\label{sec:eval_pair}

\subsection{Pseudo-pair matching} For each trimmed novice clip \(v_n^i\) with operative narration \(n^i\), we retrieve a small set of semantically similar expert clips to form \emph{pseudo-pairs} for alignment.  
Using the same sentence encoder \(f_{\text{enc}}\) applied during training data construction, we compute cosine similarity between the novice narration and all expert narrations within the same technique:
\[
s_{ij} = \cos\!\big(f_{\text{enc}}(n^i), f_{\text{enc}}(n^j_{\text{expert}})\big),
\]
and retain the top-\(k\) most similar expert clips.  
This one-to-many matching allows us to assess our editing performance against diverse variations in expert executions even within a single technique, and reduces sensitivity to outlier expert references.
Each novice clip is thus associated with $k$ candidate expert clips, denoted
\(\mathcal{P}_i = \{(v_n^i, v_{e,j}^i)\}_{j=1}^k\). We trim our expert clips using our technique-specific offsets with an additional buffer $\lambda$=30 frames \((\delta_\tau^{-} - \lambda, \delta_\tau^{+} + \lambda\))\ to allow for accurate frame-level alignment (described below).

\paragraph{Frame-level alignment.}
Because paired novice and expert motions from different actors and videos may differ in cadence and execution speed, we align each novice--expert pair using Dynamic Time Warping (DTW) on pose distance. Given two pose sequences
\[
\mathbf{p}^{\,n} \in \mathbb{R}^{T \times 3J}, 
\qquad
\mathbf{p}^{\,e} \in \mathbb{R}^{T' \times 3J},
\]
where \(J\) denotes the number of joints in the underlying pose representation, we define the framewise alignment cost as
\begin{equation}
c(\mathbf{p}^{\,n}_i, \mathbf{p}^{\,e}_j)
=
\|\mathbf{p}^{\,n}_i - \mathbf{p}^{\,e}_j\|_2^2.
\label{eq:framecost}
\end{equation}
DTW then finds the minimum-cost monotonic alignment path
\begin{equation}
\pi^*
=
\arg\min_{\pi \in \mathcal{P}}
\sum_{(i,j)\in\pi}
c(\mathbf{p}^{\,n}_i, \mathbf{p}^{\,e}_j),
\label{eq:dtw}
\end{equation}
where \(\mathcal{P}\) denotes the set of valid monotonic warping paths. The resulting alignment \(\pi^*\) defines a frame mapping \(i \mapsto \pi(i)\), which we use to resample the expert sequence to match the novice clip length \(T\).

To address left--right asymmetries between novice and expert motions (e.g., opposite striking or shooting side), we compute DTW alignment cost for both the original expert motion and its sagittally mirrored version, and retain the lower-cost alignment. This ensures that each novice motion is paired with an expert trajectory of matching chirality.

\section{Implementation details}
\label{sec:imp_det_full}

We provide further information on architecture and training hyperparameters below as mentioned in Sec. 4.1, "Implementation details."

\subsection{Pose tokenizer}

We train our pose tokenizer with $c=2$ codebooks each of size $K=256$, and latent width $d_z=256$. Following~\cite{lucas2022posegptquantizationbased3dhuman} we use a Transformer encoder with causal attention, and non-causal Transformer decoder. Both have $l=6$ layers and $n=4$ attention heads with hidden size $h_e=384$. We train for up to 1000 epochs (5000 iters/epoch) using Adam optimizer with lr=5e-5 and batch size 64.

\subsection{MotionInfiller}

MotionInfiller is a BERT-style bidirectional transformer with $l=12$ layers, hidden dimension $h_e=256$, and $n=8$ attention heads. We train with Adam optimizer using learning rate 1e-4 and batch size 16 for up to 800 epochs (5000 iters/epoch). We mask one span centered on the motion heuristic peak $t^*$. See Sec. 4.1, "Implementation details." for technique-specific motion heuristics. During training, we set maximum span mask fraction to $\alpha = 0.3$. Span lengths are uniformly sampled up to $\alpha$.

All models were trained on 8 NVIDIA Quadro RTX 6000 GPUs. Pose tokenizer training took $\sim$192 GPU hours per technique, MotionInfiller training took $\sim$96 GPU hours per technique.

\section{Training data scaling analysis}
\label{sec:scaling_analysis}

\begin{figure*}[h]
\centering
\includegraphics[width=0.9\linewidth]{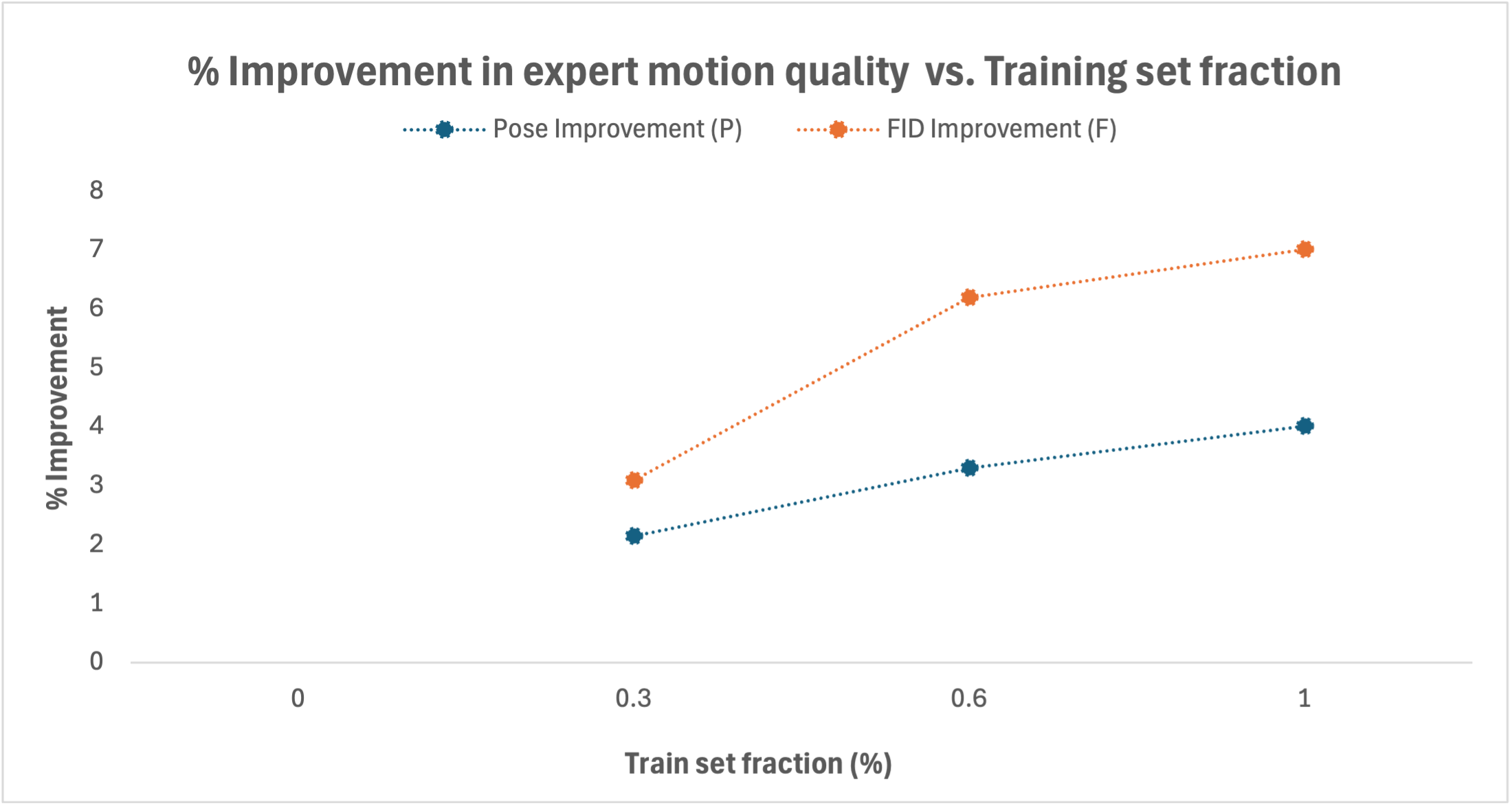}
\caption{\textbf{ExpertEdit performance as a function of training data.} We report Pose Improvement ($P$) and FID Improvement ($F$) metrics averaged across all techniques as a function of train set size. ExpertEdit performance scales well as it observes more unpaired expert video during training. } 
\label{fig:scaling_fig}
\end{figure*}

We evaluate how ExpertEdit scales with increasing quantities of unpaired expert training data by training technique-specific MotionInfiller models on 30\%, 60\%, and 100\% of the available expert clips. We report results in Fig.~\ref{fig:scaling_fig} averaged across all techniques. Both Pose Improvement ($P$) and FID Improvement ($F$) increase steadily with training set size, with especially strong gains in $F$. These results suggest that ExpertEdit continues to benefit from larger quantities of expert video, a practical advantage in our setting since training requires only unpaired expert demonstrations.

\section{Text-based motion editing prompt analysis}
\label{sec:prompt_experiments}

\setcounter{table}{2}
\begin{table*}[t]
\centering
\scriptsize
\setlength{\tabcolsep}{3.5pt}
\renewcommand{\arraystretch}{1.1}

\caption{
\textbf{Effect of different text-edit prompts on baseline motion editing performance}. We train and evaluate a representative motion-editing baseline~\cite{li2025simmotionedittextbasedhumanmotion} with different text-edit prompts. $P$ and $F$ denote relative improvement in PA-MPJPE and alignment with the expert distribution, respectively, over the source motion. Prompts encouraging general improvements in smoothness and control led to better performance than explicitly requesting expert-like motion, and adding the technique name to the prompt led to further gains.}

\begin{tabular}{l cc cc cc cc}
\toprule

& \multicolumn{8}{c}{\textbf{Basketball and Soccer}} \\

\cmidrule(lr){2-9}

& \multicolumn{2}{c}{\textbf{Mikan}}
& \multicolumn{2}{c}{\textbf{Rev. Layup}}
& \multicolumn{2}{c}{\textbf{Jumpshot}}
& \multicolumn{2}{c}{\textbf{Pen. Kick}} \\

\cmidrule(lr){2-3}
\cmidrule(lr){4-5}
\cmidrule(lr){6-7}
\cmidrule(lr){8-9}

\textbf{Prompt}
& $P(\uparrow)$ & $F(\uparrow)$
& $P(\uparrow)$ & $F(\uparrow)$
& $P(\uparrow)$ & $F(\uparrow)$
& $P(\uparrow)$ & $F(\uparrow)$ \\

\midrule

Expert-like
& 1.44 & 0.96 & 
0.97 & 0.81 & 
1.20 &  0.93 & 
1.04 &  1.12  \\

Smooth
& 2.01 & 1.11 & 
1.92 & 2.90 & 
3.12 &  1.18 & 
1.74 &  1.33  \\

Smooth + Technique
& \textbf{2.26} & \textbf{1.28} & 
\textbf{2.31} & \textbf{4.88} & 
\textbf{3.95} &  \textbf{1.24} & 
\textbf{3.20} &  \textbf{2.38}  \\

\bottomrule
\end{tabular}

\label{tab:prompt_analysis}
\end{table*}

We experimented with several motion-editing text prompts to determine which prompt is most effective for skill-driven motion editing in our baselines--without requiring personalized text corrections for each novice sample. We designed the prompts to span a meaningful range of semantic instructions: (i) explicitly requesting expert-like motion, (ii) encouraging generic improvements in motion quality (smoothness and control), and (iii) conditioning the improvement on the specific technique being performed.

\begin{itemize}
\item \textbf{Expert-like:} ``Make the motion look more expert-like in form.''

\item \textbf{Smooth:} ``Make the motion smoother and more controlled.''

\item \textbf{Smooth + Technique:} ``Make the \texttt{<TECHNIQUE\_NAME>} motion smoother and more controlled,'' where \texttt{<TECHNIQUE\_NAME>} is replaced with the human-readable technique name (e.g., ``reverse layup,'' ``mid-range jumpshot,'' ``roundhouse kick'').
\end{itemize}

We evaluate these prompt variants using SimMotionEdit~\cite{li2025simmotionedittextbasedhumanmotion}, a representative text-driven motion editing baseline described in Sec.~4.1, and show results in Table~\ref{tab:prompt_analysis}. We find that including ``expert-like" motion in the prompt leads to poor performance, likely because the datasets our baseline models were pre-trained on (e.g., MotionFix~\cite{athanasiou2024motionfixtextdriven3dhuman}) do not capture text edits related to expertise. Instead, encouraging motion to become ``smoother and more controlled" leads to improved performance on our expert-like motion quality metrics. Furthermore, adding the technique name to the prompt leads to the best performance. We therefore report results with the \textbf{Smooth + Technique} prompt for all baselines in Tables~1 and~2.

\section{Mask span length ablation}
\label{sec:mask_span_ablation}
We ablate the maximum mask span fraction $\alpha$, which controls how much of the skill-critical phase is infilled. Results averaged across techniques are shown in Table~\ref{tab:alpha_ablation}. A smaller span slightly improves pose error, while the default $\alpha=0.3$ gives substantially stronger distributional alignment to expert motion.

\begin{table}[h]
\centering
\small
\caption{\textbf{Mask-span ablation.} Average Pose Improvement ($P$) and FID Improvement ($F$) across techniques.}
\label{tab:alpha_ablation}
\begin{tabular}{lcc}
\toprule
Variant & Avg. $P\uparrow$ & Avg. $F\uparrow$ \\
\midrule
$\alpha=0.15$ & \textbf{4.09} & 6.10 \\
$\alpha=0.3$ \textbf{(default)} & 4.02 & \textbf{7.01} \\
$\alpha=0.45$ & 3.88 & 6.41 \\
\bottomrule
\end{tabular}
\end{table}

\section{Absolute metric values}
\label{sec:absolute_metrics_sec}

In the main paper, we report relative Pose Improvement ($P$) and FID Improvement ($F$)
to normalize improvement across datasets, skeletons, and techniques. Here, we provide
the corresponding raw PA-MPJPE and FID values for scale. Table~\ref{tab:absolute_metrics}
reports averages across all eight techniques for the source novice motions, ExpertEdit,
and SimMotionEdit, the strongest baseline evaluated across all techniques. ExpertEdit
achieves the lowest absolute PA-MPJPE and FID, preserving the same ranking as the
relative metrics.

\begin{table}[h]
\centering
\small
\caption{\textbf{Absolute metrics, averaged across techniques vs. strongest baseline SimMotionEdit.}
ExpertEdit achieves the lowest raw PA-MPJPE and FID, confirming that the relative
improvement metrics in the main paper are consistent with absolute metric values.}
\label{tab:absolute_metrics}
\begin{tabular}{lcc}
\toprule
Method & PA-MPJPE$\downarrow$ & FID$\downarrow$ \\
\midrule
Source novice & 204.0 & 220.5 \\
SimMotionEdit & 198.9 & 213.5 \\
ExpertEdit & \textbf{195.8} & \textbf{205.1} \\
\bottomrule
\end{tabular}
\end{table}

\section{Qualitative results video}
\label{sec:supp_video}

See the project page for a video containing qualitative examples of human-verified pseudo-pairs, motion editing results across techniques, and a representative failure case.
\end{document}